\documentclass[10pt,letterpaper]{article}

\usepackage{cogsci}
\usepackage{pslatex}
\usepackage{apacite}
\usepackage{url}
\title{Explaining Necessary Truths}
 \author{Gülce Kardeş \\
            Department of Computer Science, University of Colorado, Boulder, CO 80309, USA \\ \& the Santa Fe Institute, Santa Fe, NM 87501, USA  
\\ {\bf Simon DeDeo} \\
    Carnegie Mellon University, Pittsburgh, PA 15213, USA \\ \& the Santa Fe Institute, Santa Fe, NM 87501, USA \\
    \texttt{sdedeo@andrew.cmu.edu}
 }
\cogscifinaltrue

\begin{document}

\maketitle

\begin{abstract}
Knowing the truth is rarely enough---we also seek out reasons \emph{why} the fact is true. While much is known about how we explain contingent truths, we understand less about how we explain facts, such as those in mathematics, that are true as a matter of logical necessity. We present a framework, based in computational complexity, where explanations for deductive truths co-emerge with discoveries of simplifying steps during the search process. When such structures are missing, we revert, in turn, to error-based reasons, where a (corrected) mistake can serve as fictitious, but explanatory, contingency-cause: not making the mistake serves as a reason why the truth takes the form it does. We simulate human subjects, using GPT-4o, presented with {\bf SAT} puzzles of varying complexity and reasonableness, validating our theory and showing how its predictions can be tested in future human studies. \\
\textbf{Keywords: explanation; deduction; logic; mathematical cognition; proofs; reasoning; computational complexity} 
\end{abstract}

\noindent Explanations---the “reasons why” something is true---often refer to the contingencies of life. What we identify, for example, as the reason why we passed an exam, will often refer to things that could have gone differently: we passed the exam, for example, because we studied hard (but might not have), because the class was easy (but might have been hard), or because we got enough sleep the night before (but might have stayed up). When explanations are hard to come by, conversely, we often look for hidden contingencies so that ``why" questions lead us to seek out new information. A growing literature, centered around Bayesian, probabilistic accounts of how contingencies compose together, tells us, in turn, how to select between different models on the basis of both their raw likelihood, and value judgements such as simplicity, unification, or consilience~\cite{wojtowicz2020probability}.

Not all explanations, however, refer to things that could have gone otherwise. In many cases, the challenge is not in pointing to a particular contingency, but in showing how (as a deductive matter, given the models to hand) the different contingencies interact with each other to produce the thing to be explained. There is a computational challenge implicit in explanatory work, in addition to the challenge of gathering information; as noted by \citeA{roth1996hardness}, Bayesian reasoning involves counting up the different ways contingencies interact, a problem whose general solution is in the computational complexity class {\bf \#P} (``sharp P'', even more resource-demanding than {\bf NP}) and involves, potentially, mentally simulating an exponentially-branching number of possibilities.

It is natural, in other words, to think that explanations refer not only to the contingencies at the base of different models, but also to the logical form of the models themselves, where the challenge, for computationally-limited beings, is not in the search for new information, but in the task of thinking things through. A good explanation can, for example, reveal a hidden property in a model, that the recipient already possesses, which makes its consequences easier to see. While it may be easy to see how studying leads to good exam performance, it might be much harder to see how different theories about a patient's medical conditions make different diseases more or less likely.

The role of explanations as computational aids is most obvious in the case of purely deductive reasoning, such as occurs in mathematical proofs. If we accept that it makes sense to ask ``why'' a theorem is true, we are faced with the problem that there are no contingencies at all: once we believe the axioms \cite{maddy1988believing}, there is nothing, on pain of logical inconsistency, that could have gone the other way.

In simple cases, simply repeating the axioms may be enough. Mathematical deductions, however, may begin with a small number of axioms, but quickly ramify: as shown by \citeA{viteri2022epistemic}, a formal statement of a mathematical theorem involves tens of thousands of logical steps, and even human-readable proofs stretch to hundreds of pages---while mathematicians talk freely about the reasons why a theorem or conjecture is true \cite{gowers2023makes} that go well beyond a simple reassertion of the necessary definitions. Necessary truths emerge, just as much, in explaining the output of complex computations that occur in neural networks---as does the need to provide explanations for them, an active area for current research in computer science \cite{barcelo2025explaining}.

To study the explanation problem, we draw on recent work that connects basic questions in cognitive science to computational complexity theory (CCT). CCT is already the basis of recent exciting empirical work in error-making and difficulty~\cite{exp1,exp3,exp4,exp5,more_exp}, in theory~\cite{iris08,van2019cognition} (the tractable cognition thesis), and for questions in AI~\cite{irisAI}, economics~\cite{econ}, and evolution~\cite{iris, rich2021naturalism}.

We focus in particular on explanations for what are called {\bf SAT} or ``satisfiabilityproblems ''. In a {\bf SAT} problem, the subject is presented with a logical formula, and asked to find an assignment to the variables that makes the formula true; for example, if the two variables are $x_1$ and $x_2$, then the following formula,
\begin{equation}
(x_1) \land (x_2 \lor \lnot x_1),
\label{easy}
\end{equation}
or, in words, ``$x_1$ is true, and either $x_2$ is true, or $x_1$ is false'', has only one solution: $x_1$ and $x_2$ both set to true. Eq.~\ref{easy} takes the ``canonical'' form for a {\bf SAT} problem, known as conjunctive normal form (CNF), where a set of clauses containing only ``or''s is composed together with ``and''s; CNF is a particularly useful form for these problems because by varying simple properties (\emph{e.g.}, the length of each clause, or the number of times the same variable appears in different clauses), problems can be made increasingly hard. When formulas contain clauses with three or more variables (known as $k$-$\mathbf{SAT}$, with $\mathbf{3}$-$\mathbf{SAT}$ being a special case), the difficulty of finding solutions can vary dramatically based on the structure of the formula. Formulas where variables appear repeatedly across different clauses in certain patterns can be particularly challenging in practice \cite{achlioptas2006random}. For these instances, even our most advanced algorithms typically require exponential time to determine an assignment of true/false values that satisfies all clauses.

Although CCT results often talk about how hardness  emerges in the infinite limit \cite{Moore2011}, the cognitive challenges emerge early. Eq.~\ref{easy} is trivial (it is, in fact, a rephrasing of $x_1$ and $x_1 \rightarrow x_2$), but a few more clauses suffice to make things far more opaque; consider, for example,
\begin{eqnarray}
(\lnot x_1 \lor \lnot x_2 \lor \lnot x_3 \lor \lnot x_4) \land (\lnot x_1 \lor \lnot x_2 \lor x_4) \land \nonumber \\
(x_1 \lor \lnot x_3) \land (x_2 \lor \lnot x_3 \lor \lnot x_4) \land (x_3 \lor \lnot x_4) \land (x_3 \lor x_4)
\label{hard}
\end{eqnarray}
Eq.~\ref{hard} has, like Eq.~\ref{easy}, only one solution ($x_1$ and $x_3$ true; $x_2$ and $x_4$ false---{\tt TFTF}); like Eq.~\ref{easy}, each clause is necessary for that solution to be the only one; but, unlike Eq.~\ref{easy}, it is not immediately clear how the solution emerges from the constraints.

\section{Kinds of Reasons Why}

{\bf SAT} instances like Eq.~\ref{hard} form the test-bed of this paper's study of how explanation works in the absence of contingency; given the axioms of logic, all six clauses, and all four variables, contribute equally. And yet, at the same time, we can see how particular reasons why might emerge: looking at the final pair of clauses in Eq.~\ref{hard}, for example, we can see that $x_3$ has to be true---which (clause 3) implies that $x_1$ has to be true, which simplifies clauses one and two into $(\lnot x_2 \lor \lnot x_4)\land(\lnot x_2 \lor x_4)$, which implies, by a similar pattern, that $x_2$ has to be false, and finally (clause 4) that $x_4$ must be false.

Our deduction, in other words, proceeded $x_3\rightarrow x_1 \rightarrow \lnot x_2 \rightarrow \lnot x_4$; informally, a cognitively-limited agent would prefer to say that the truth of $x_3$, rather than the falsehood of $x_4$, is a better reason why the full solution is {\tt TFTF}---even though a deductive proof could proceed with any variable at all; beginning (for example) with the falsehood of $x_4$ to deduce the truth of $x_3$, and so forth. In this case, the initial simplification that enabled us to derive $x_3$ is known as a ``resolution''; another obvious heuristic is to work with clauses that contain only one variable (so-called ``unit clauses'')---these serve as simple axioms. Beginning with $\lnot x_4$ in Eq.~\ref{hard} would appear to require a lucky, or perhaps ``intuitive'', guess.

While heuristics like resolution can simplify some problems, they can not, in general, solve all of them. In computer science, the earliest \textbf{SAT} solvers such as the Davis-Putnam procedure work step by step, choosing a variable at random, assigning an arbitrary truth value and propagating forward the consequences until either a solution, or a contradiction, is reached---in the latter case, the procedure ``backtracks'' to a previous step and tries the opposite assignment, repeatedly going forward and backward until a solution is found. Later variants uses heuristics to improve efficiency, focusing on (for example) variables that appear most frequently in the formula, as these often have the largest impact on satisfiability \cite{MarquesSilva1999}; this captures a second style of reason why, that focuses on the influence a variable has on different parts of the problem: a variable that appears more often provides more reasons for the others to take the values they do.

Modulo the possession of a special faculty of insight unavailable to machines \cite{parsons1995platonism,penrose2016emperor} the same limitations of these industrial {\bf SAT} solvers ought to apply to humans, and {\bf SAT} problems should be full of ``unreasonable'' truths \cite{dedeo2024hard}, where efficient simplifications are hard, or impossible, to find. In these cases, depth-first search with backtracking, in the Davis-Putnam style, with or without heuristics, is a natural tactic.  

In the absence of simplifications, backtracking can provide a novel source of reasons why distinct from simplification: rather than explaining an answer by reference to a simplifying structural feature of the problem, one can do so by reference to a mistake that one ought not to make. If you begin trying to solve Eq.~\ref{hard} by picking $x_4$ and seeing what happens if it is set to ``true'', you will eventually reach a contradiction that will require you to return to $x_4$ and set it to ``false'' instead. In this case, one can explain the final answer in terms of the better choice---a form of fictitious cause, grounded not in logic, but the contingent sequence of choices made by the solver.

\section{Computational Experiments}

The {\bf SAT} instances that probe the kind of explanation-making we care about are challenging and potentially wearying to solve, and there is a lot we do not understand about how to best construct a psychological experiment to test these predictions in actual human subjects. Large-language models such as GPT-4o show a great deal of promise as a tool to approximate human behavior at a level sufficient to serve as proof of concept for more expensive, smaller-scale studies of human behavior. In general, LLMs appear to replicate the kinds of reasons ordinary subject might give; the reasons given by LLMs are also of intrinsic interest, of course, whether or not they deviate from those that might be provided by more or less distracted humans. We emphasise that these experiments do not count as evidence in favor of actual human performance, but rather as proof of concept that our frameworks above are, indeed, both testable and falsifiable.

Our prompt for our computational experiments, fed via OpenAI's API to GPT-4o, is 
\begin{quote}
Here's a SAT formula.

[formula]

Talk through the finding a solution for this SAT formula.

Once you think you have a solution, double check it to make sure that it's correct. If not, keep reasoning to get the answer, and if you get a new one, double check it as well, and keep double-checking carefully until you think you have the answer. Keep track of any assumptions that you make that later turn out to be false.

Then, at the end of your talking, tell me the main reason \emph{why} this is the solution, focusing on a single variable. Do not use any python code or outside tools.

Return, at the end of your response, a JSON object, with four fields. The first field, SOLUTION, should be a string with only T and F providing the satisfying assignment in order. The second field, REASON, should be an integer from 1 to 4, giving the name of the variable that is the main reason \emph{why} this is the solution. The third field, EXPLANATION, should be a string that contains your explanation why this is a solution. If you made an assumption that later turned out to be false, the fourth field, ERROR, should contain the integer name of the variable you made the incorrect assumption for, and -1 otherwise.
\end{quote}

As test cases, we consider {\bf SAT} problems with four variables and between four and six clauses. We require that each problem have a single unique solution, and that every clause and every variable matter (\emph{i.e.}, if any clause is deleted, the number of solutions increases). To study the impact of simplifying heuristics on given reasons, we consider a range of problem complexity, ranging from problems containing a single unit clause (\emph{i.e.}, a clause that fixes one variable as an axiom) to problems containing a simple resolution pair and, finally, to problems with neither unit clauses nor simple resolution pairs. We consider 400 distinct problems of each type, and provide the system with 20 version of each problem with the order of clauses and variables shuffled. The total cost for these 32,000 runs was approximately 1000 USD.

\section{Results: Identifying Reasons}

\begin{table*}[!h]
    \centering
    \begin{tabular}{c|c|c||c|c|c}
    Reason & Used When  & Used When & Competing Simplification &  Competing Backtrack & Influence \\
    & Possible? & Needed? & (logistic coeff.) & (logistic coeff.) & (logistic coeff.) \\ \hline
    Unit Clause & 52\% & 68\% & $-0.49\pm0.03^{\star\star\star}$ & $-0.65\pm0.04^{\star\star\star}$ & $+1.39\pm0.05^{\star\star\star}$\\
    Resolution & 35\% & 50\% & $-1.04\pm0.04^{\star\star\star}$ & $-0.21\pm0.04^{\star\star\star}$ & $+1.41\pm0.04^{\star\star\star}$ \\ 
    Backtrack & 36\% & 38\% & $-0.25\pm0.05^{\star\star}$ & --- & $+1.46\pm0.05^{\star\star\star}$ \\
    \end{tabular}
    \caption{Shortcuts and simplifications can serve as reasons why: simulated subjects cite the variable in a unit clause, or the outcome of a resolution, as a reason why at rates well above chance, with unit clause reasons being the strongest, and outcompeting resolution when both are present. These reasons why, in turn, can be displaced when a deduction leads to a contradiction and backtracking. Finally, when a reason can be associated with a variable's repeated appearance in multiple clauses (influence), it is more likely to be cited. In this simulated study, $N$=31,988; $^{\star\star\star}$ indicates $p\ll10^{-3}$; $^{\star\star}$, $p<10^{-2}$.}
    \label{drivers}
\end{table*}
On the basis of the discussion above, we propose, and test, three main hypotheses for how simulated reasoners identify the reasons behind a logical truth. We track how often a reason is identified when it is possible to do so (\emph{i.e.}, when the simplification exists, does the subject identify the hinge variable of that simplification as the reason why, or when backtracking has been done, does the subject identify the backtracked assumption variable as the reason why). We also track the role that ``influence''---the repeated appearance of a variable across multiple clauses---plays in shifting the location of the identified reason; while influence, alone, does not provide a direct mechanism for either simplification or error-correction, a focus on these high-degree variables can provide cognitively-efficient ways to understand how multiple constraints interact. A simple multiple-variable logistic regression helps us determine what might affect the deployment of a reason depending on the presence of competing reasons, including these high-influence variables that connect multiple parts of the problem space. We have three main hypotheses.

{\bf H1}. In the presence of a clear simplification, the implicated variable is more likely to be cited as a reason why. We find this to be the case for both the unit clause case (the ``axiomatic'' variable is cited as a reason 52\% of the time, and 68\% of the time when no competing reason exists; $p\ll 10^{-3}$), and the resolution case (when a necessary truth exists in a resolution pair, the relevant variable is identified as a reason 34\% of the time, and 50\% of the time when no competing reason exists; $p\ll 10^{-3}$). Indeed, the two reasons interfere; the presence of a unit clause reduces the likelihood of a resolution explanation, and vice versa ($p\ll 10^{-3}$; Table~\ref{drivers}).

{\bf H2}. Backtracking, when it exists, can serve as a reason for deductive truth through the fictitious contingency effect. We find this to be the case; when the solution was found in a process that included backtracking, the backtracked variable was cited as the reason 36\% of the time, and 38\% when only backtracking was present; $p\ll 10^{-3}$. Backtracking also directs attention away from simplifications: as shown in Table~\ref{drivers}, when a deduction leads one down a blind alley, the ``blame'' for the bad branch competes with the simplification as a reason why.

{\bf H3}. Influence reasons lend support to both simplifying and fictitious causes; when a simplifying or a backtracked variable also happens to have maximum degree, it is more likely to be identified as a cause. Again, as shown in Table~\ref{drivers}, this influence effect is a strong driver of the identification.

\section{Results: Describing Reasons}

\begin{table*}[!h]
    \centering
    \begin{tabular}{c|c||c|c|c|c}
    Language & Baseline & Unit Clause & Resolution & Influence & Backtrack \\
    & Frequency & (logistic coeff.) & (logistic coeff.) & (logistic coeff.) & (logistic coeff.) \\ \hline
    Causation & 25\% & $+0.95\pm0.3^{\star\star\star}$ & (n.d.) & (n.d.) & (n.d) \\
    Simplification & 29\% & $0.60\pm0.03^{\star\star\star}$ & $+0.11\pm0.03^{\star\star}$ & $+0.20\pm0.03^{\star\star\star}$ & $-0.92\pm0.06^{\star\star\star}$ \\
    Importance & 42\% & $-0.46\pm0.03^{\star\star\star}$ & (n.d.) & $+0.5\pm0.03^{\star\star\star}$ & $-0.57\pm0.04^{\star\star\star}$\\ 
    Counterfactual & 1.7\% & $-1.5\pm0.2^{\star\star\star}$ & $-0.4\pm0.2^{\star}$ & $-0.3\pm0.1^{\star}$ & $+0.8\pm0.2^{\star\star\star}$ \\
    Contradiction & 22\% & $-0.27\pm0.04^{\star\star\star}$ & (n.d.) & (n.d.) & $+0.54\pm0.05^{\star\star\star}$ \\
    \end{tabular}
    \caption{Linguistic signals of reason-giving: simplification language is associated with simplification reasons, as expected, and anti-associated with backtracking. Conversely, counterfactual talk (rare) and appeals to logical consistency or contradiction is associated with backtracking. Talk about importance and influence is more likely when the reason-why variable has maximum degree. Interestingly, strong causal signals appear only for unit clauses: (simple) axioms are described as causes. {\bf Causation} words: forc*, require*, impact*, relies, fixes, constrains, caus*, effect*, dictate*. {\bf Simplification} words: simpli*, easier, key. {\bf Importance} words: mult*, strong, importan*, pivotal, crucial, critic*, central, influential, hinge, vital. {\bf Counterfactual} words: otherwise, if, would, could, should, unless, instead, although, despite. {\bf Contradiction} words: only, contradict*, necess*, consisten*.}
    \label{words}
\end{table*}

The previous section considered the ways in which a particular variable was identified as a reason why, and how this was predicted by the roles it can play in the underlying search and reasoning process. In addition to asking simulated subjects for raw identification, we also had them provide natural-language accounts of why the variable they identified served as a reason why, and we then analyzed that language with simple word count heuristics.

Table~\ref{words} shows the results. As expected, when variables that can play a role in a simplification are identified as reasons why, the associated language refers to how they make things easier, simplify things, or how the identification is a ``key'' step in finding a solution. Unit clauses, and high-degree variables, are associated with talk about their ``critical'' or ``pivotal'' features. 

Intriguingly, blatant causation talk is primarily associated with unit clause identification: these simple clauses, amounting to basic axioms of the theory, take on a causal tone, while backtracking identification, by contrast, is associated with talk in terms of necessity and the avoidance of contradiction. Very little talk has standard signals of counterfactuality such as ``would'', ``should'', and ``otherwise'', although these are, as expected, strongly associated with backtrack reasons alone.

\section{Discussion}

Our main goal in this paper is to draw attention to a puzzling phenomenon: even in cases where all the facts are networked together in logical (rather than causal) patterns, we still seek, and talk about, reasons why these facts hold. This happens even in cases where, as a matter of logic, the rules of the system mean that any truth can be considered primary: if one is given $A$ and $\lnot A \lor B$, and discovers (somehow) that this requires that both $A$ and $B$ are true, one can just as easily say that A is true because B is true (\emph{modus tollens}) as that B is true because A is true (\emph{modus ponens}).

We have presented, in turn, an account of necessary explanation that hinges on two key cognitive phenomena. First, the drive for sense-making and simplification \cite{chater2016under}, well-known to act in causal-contingent explanatory practices \cite{lombrozo2007simplicity}, ought to lead subjects to consider moves that simplify the search process---unit clauses corresponding to axioms, or resolution-style reasons that hinge on $A$ being a consequence of both $B$ and $\lnot B$---as reasons for the final truth of the matter.

Second, we show how reasons for why can emerge from the contingent failures that arise when solving unfamiliar problems with limited cognitive resources. When agents fail to find powerful simplifying reasons—an inevitable occurrence given that logical satisfiability is \textbf{NP}-complete—they must backtrack, revising their initial assumptions to avoid logical contradictions. This backtracking process leads to what we call "fictitious contingency" explanations: variables whose values needed adjustment during search are more likely to be cited as reasons for the final solution. These backtrack-induced reasons take a contrapositive form: “X must be true because setting it false leads to a chain of deductions that end in contradiction.” The prominence of X in the explanation stems not from any inherent logical priority, but merely from the path our search process happened to take. As we have seen (Table~\ref{drivers}) these fictitious contingencies can emerge even when better reasons are available.

While it is well known that simplicity exercises a fascination for the human mind, the role of error-making in explanation is less familiar. In contrast to the errors we make in predicting contingent facts, the errors we make in logic ought to take us nowhere---by the principle of explosion, at the very least, if we assume a contradiction, we can derive everything and nothing \cite{munroe}, and the emergence of these trivialities itself can serve as a heuristic detector of error \cite{dedeo2024alephzero,olga}.

Accounts by practicing mathematicians, however, argue otherwise. The mathematician Alexander Grothendieck, for example, refers to the ``repeated knocking and probing'' that mathematicians undertake~\cite{alex}, that leads to ``the detection of a false idea from the first ``break-offs'' observed between the image obtained and certain obvious facts'' and which makes the discovery of  error ``one of the crucial moments, a creative moment among all, in all works of discovery''. If errors are creative processes that both create and refine the reasoner's cognitive model, our framework provides an account of how such errors enable a novel form of reasoning based on the contingencies of exploration.

Our results also point to further sources of logical explanation and new places of contact between the cognitive and computer sciences. Modern {\bf SAT} solvers, for example, employ conflict-driven clause learning (CDCL; \citeA{marques2009conflict}), which analyzes failed assignment attempts to derive new clauses that prevent similar failures ----a potential analogy for the kind of error-based reasoning described by Grothendieck. However, even these sophisticated approaches can struggle with the hardest instances, which typically lack unit clauses and simple resolutions, and exhibit, instead, the kind of high uniformity---where variables appear together with similar frequencies across clauses, with no single variable having outsized impact---that frustrates influence-based reasons as well.

CDCL algorithms struggle with problems that possess high amounts of symmetry where permuting variables preserves the formula's overall structure. Symmetry makes problems particularly challenging because different variable assignments can be structurally equivalent and lead to redundantly unproductive, but hard to spot, search paths where the solver cannot leverage learned information from one branch to prune others. This balanced structure leaves solvers without clear signals about which assignments to try first. We expect similar effects to obtain for human reasoners, with knock-on effects for both the reasons they give to explain the solution, and the extent to which they consider the facts, lost as they are in a sea of uniformity, explainable at all.

This challenge of navigating symmetries and contradictions opens up broader questions about how discoveries in logio-deductive systems, such as the discovery of proofs, emerge: encountering a contradiction after several deductive steps---the essence of our proposed backtracking-reason mechanism---can often reveal a critical branch point in the structure, which is pointed to by a prior immediately prior decision point. These bifurcation points likely play different roles in expert versus novice proof construction and problem-solving, suggesting a key dimension along which mathematical maturity develops: more sophisticated reasoners may be able to both project forward further, identifying branch points more efficiently, and to recall the prior decision-history more easily, making backtracking more efficient. 

Our work with \textbf{SAT} shows how these contradictions serve as reasons-why that can complement those found in simplification discoveries, but the relationship between expertise and contradiction-handling remains unexplored. Future studies could investigate whether experts show systematic differences in how they leverage contradictions, moving beyond local navigation via backtracking to extract broader structural insights about the proof space.

Empirically validating the reasoning patterns and explanatory strategies discussed in this paper is challenging, because they are expected to emerge only when subjects engage, over an extended period, with a difficult problem, and master the relevant notation, heuristics, and (potentially) cover stories. To help refine our ideas, we conducted experiments with simulated subjects; these experiments lend credence to the idea that these systematic explanation biases should be detectable in real-world experiments. These pilot results not only give us insight into the reasoning methods of automated systems but also now enable us to plan a series of experiments at our home institutions involving undergraduates in mathematics, computer science, and logic.

For many researchers, a better understanding of the purely deductive explanation is valuable in its own right: sophisticated mathematical-deductive cognition may be evolutionarily novel, but it is culturally ancient, appearing at approximately the same time as philosophy, tragedy and the theological innovations of the Axial Age \cite{bellah2011religion}.

The question of deductive explanation, however, reaches well beyond the provinces of mathematical proof. As noted in the introduction to this piece, even when we reason about contingencies, we are usually faced with models of sufficient complexity as to pose calculational challenges equal to, if not exceeding, those of information-gathering. Using Bayes' rule requires us to assess prior probabilities, and to estimate the conditional probabilities of observations given a theory to hand---but it also can require us, in computing the denominator, to solve challenging problems that are no less deductive than {\bf SAT} problem like those of Eq.~\ref{hard}. Explaining why we believe this or that theory is the better one to believe may often, in turn, focus less on the evidence in the world, and much more on the kinds of simplicity and backtracking-induced reasons that a {\bf SAT}-style investigation lays bare.




\noindent {\bf Acknowledgements}. SD acknowledges the support of the Survival and Flourishing Fund. GK acknowledges the support of a Santa Fe Institute Graduate Fellowship, the John Templeton Foundation, and J.\ Grochow and R.\ Frongillo Startup Funds at the University of Colorado Boulder.

\bibliographystyle{apacite}

\setlength{\bibleftmargin}{.125in}
\setlength{\bibindent}{-\bibleftmargin}

\bibliography{references}

\end{document}